\pdfoutput=1

\documentclass[10pt,twocolumn,letterpaper]{article}

\usepackage{cvpr}

\usepackage[pagebackref,breaklinks,colorlinks,allcolors=blue]{hyperref}

\usepackage{times}
\usepackage{latexsym}

\usepackage[T1]{fontenc}
\usepackage{arabtex}
\usepackage{utf8}
\setcode{utf8}

\usepackage[utf8]{inputenc}

\usepackage{microtype}
\usepackage{graphicx}   
\usepackage{caption}    
\usepackage{multirow}
\usepackage{tabularx}
\usepackage{booktabs}
\definecolor{citecolor}{HTML}{0071BC}
\definecolor{linkcolor}{HTML}{ED1C24}
\definecolor{LGray}{gray}{0.97}
\usepackage{multicol}
\usepackage{colortbl}
\usepackage{xcolor}
\definecolor{tabhighlight}{HTML}{e5e5e5}
\usepackage{color}
\usepackage{xspace}
\usepackage{pifont}
\usepackage{colortbl}

\definecolor{darkgreen}{rgb}{0.0, 0.5, 0.0}  
\newcommand{\cmark}{\textcolor{darkgreen}{\scalebox{1}[1.0]{\ding{51}}}}
\newcommand{\xmark}{\textcolor{red}{\ding{55}}}  

\title{
  {\includegraphics[width=1cm]{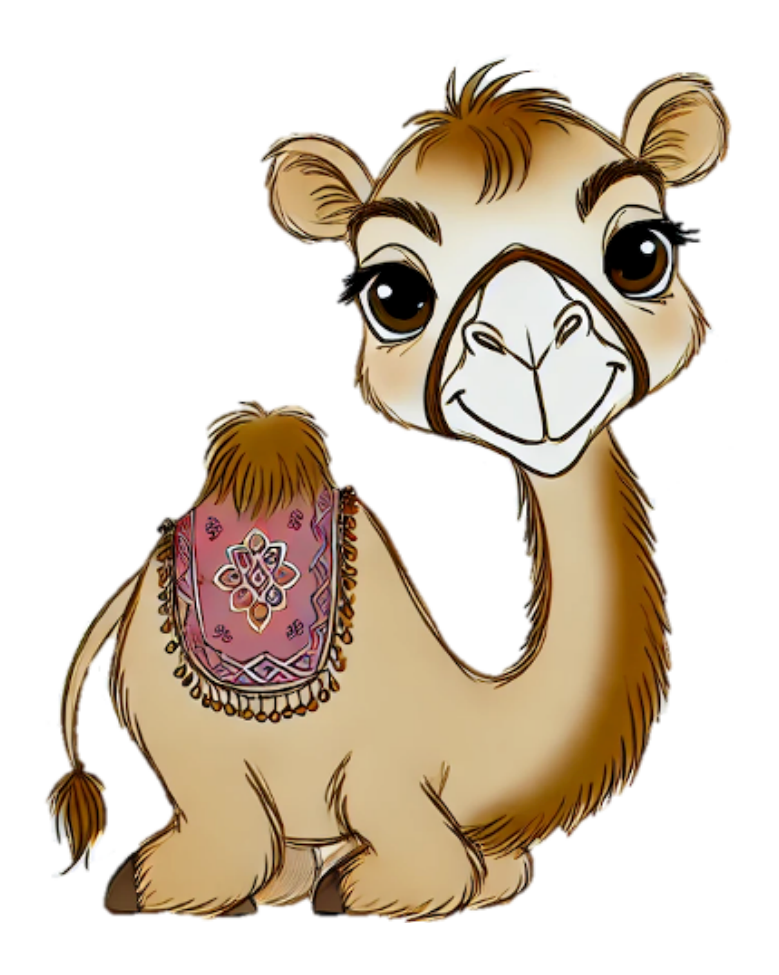} CAMEL-Bench: A Comprehensive Arabic LMM  Benchmark
}}

\begin{document}

\author{\\
    {Sara Ghaboura}\textsuperscript{1*} \quad {Ahmed Heakl}\textsuperscript{1*} \quad {Omkar Thawakar}\textsuperscript{1} \quad {Ali Alharthi}\textsuperscript{1} \\
     {Ines Riahi}\textsuperscript{2} \quad  {Abduljalil Saif}\textsuperscript{2} \quad  {Jorma Laaksonen}\textsuperscript{2} \quad  {Fahad Shahbaz Khan}\textsuperscript{1,3} \\
     {Salman Khan}\textsuperscript{1,4} \quad  {Rao Muhammad Anwer}\textsuperscript{1,2} \\[0.5em]
    {\fontsize{10.5pt}{12pt}\selectfont \textsuperscript{1}Mohamed bin Zayed University of AI, \textsuperscript{2}Aalto University, \textsuperscript{3}Linköping University, \textsuperscript{4}Australian National University
    }\\
    {\hypersetup{urlcolor=blue}
  \fontsize{11pt}{12pt}\selectfont \href{https://mbzuai-oryx.github.io/Camel-Bench/}{https://mbzuai-oryx.github.io/Camel-Bench/}}
    }

\maketitle

\begin{abstract}
Recent years have witnessed a significant interest in developing large multimodal models (LMMs) capable of performing various visual reasoning and understanding tasks. This has led to the introduction of multiple LMM benchmarks to evaluate LMMs on different tasks. However, most existing LMM evaluation benchmarks are predominantly English-centric. In this work, we develop a comprehensive LMM evaluation benchmark for the Arabic language to represent a large population of over 400 million speakers. The proposed benchmark, named CAMEL-Bench, comprises eight diverse domains and 38 sub-domains including, multi-image understanding, complex visual perception, handwritten document understanding, video understanding, medical imaging, plant diseases, and remote sensing-based land use understanding to evaluate broad scenario generalizability. Our CAMEL-Bench comprises around 29,036 questions that are filtered from a larger pool of samples, where the quality is manually verified by native speakers to ensure reliable model assessment. We conduct evaluations of both closed-source, including GPT-4 series, and open-source LMMs. Our analysis reveals the need for substantial improvement, especially among the best open-source models, with even the closed-source GPT-4o  achieving an overall score of 62$\%$. Our benchmark and evaluation scripts are open-sourced.\footnote{* Equal Contributions} 
\end{abstract}

\begin{figure}[t!]
    \centering
        \includegraphics[width=\columnwidth]{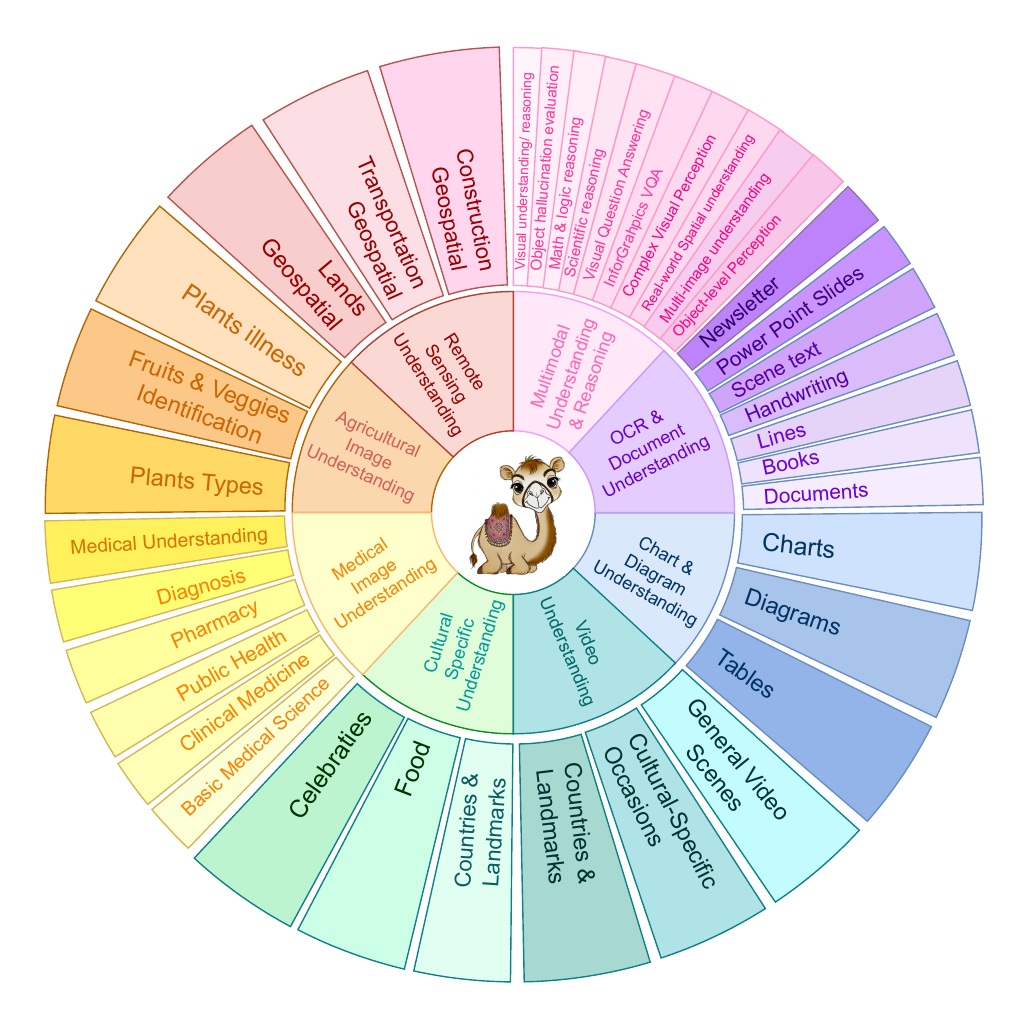}\\
        \caption{The proposed CAMEL-Bench covers eight diverse and challenging domains: \textit{multimodal understanding and reasoning}, \textit{OCR and}  \textit{documents}, \textit{charts and} \textit{diagrams}, \textit{videos}, \textit{cultural-specific content}, \textit{medical images}, \textit{agricultural images}, and \textit{remote sensing understanding} in Arabic. CAMEL-Bench covers 38 sub-domains with over 29K questions carefully curated by native Arabic speakers to rigorously evaluate essential skills desired in Arabic LMMs.}
        \label{fig1_hierarchy}
        \vspace{-1em}
\end{figure}

\section{Introduction}
Large multimodal models (LMMs) have recently achieved significant advancements across a broad spectrum of tasks, including visual reasoning, perception, and multimodal understanding. Closed-source models such as GPT-4V and open-source LMMs, such as LLaVA \cite{liu2023llava} have demonstrated effectiveness in tasks like image captioning \cite{radford2021learning}, visual question answering (VQA) \cite{li2022blip,li2023blip}, and complex visual reasoning \cite{cho2021vlt5}.
These recent developments have led to the introduction of different benchmarks to evaluate the performance of open and closed-source LMMs. Despite these advances, most existing LMM benchmarks are English-centric, limiting their applicability to other languages \cite{PALO}.

\begin{figure*}[!ht]
    \centering
      \includegraphics[width=1\textwidth]{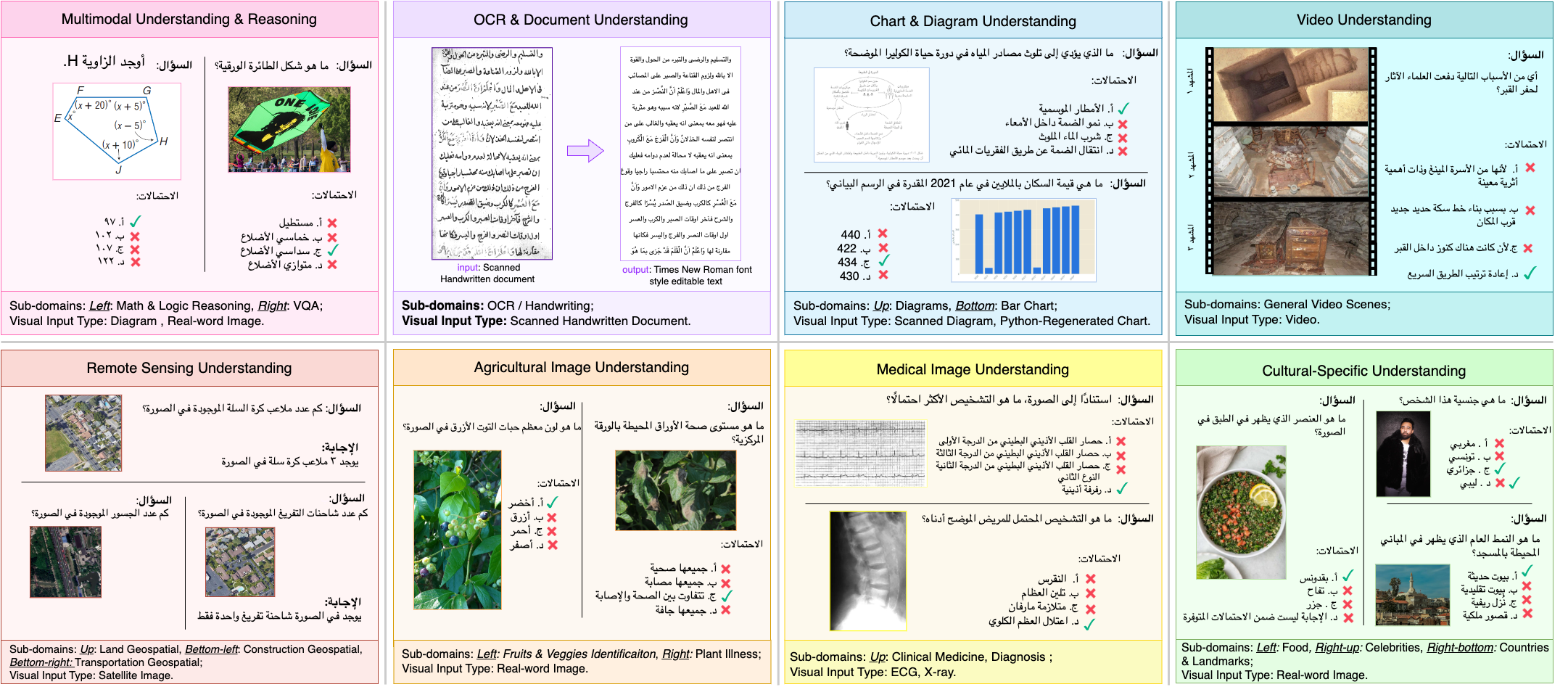}
        \caption{CAMEL-Bench examples spanning eight diverse domains, encompassing a wide range of visual data types and tasks.}
        \label{fig:qual_samples}
\end{figure*}

With over 400 million speakers, Arabic is the 5\textsuperscript{th} most widely spoken languages globally. In the context of large language models (LLMs), there exist various attempts in developing Arabic LLMs \cite{sengupta2023jais, huang2023acegpt} which has also led to the introduction of Arabic LLM benchmarks \cite{koto2024arabicmmlu}. In the context of LMMs, few recent works explore Arabic-centric evaluations in certain areas such as, scientific exams \cite{das2024exams}, cultural aspects \cite{romero2024cvqa, alwajih2024peacock}, Arabic question answers and documents \cite{abdallah2024arabicaqa, mahmoud2014khatt}. However, there is still a lack of comprehensive and diverse Arabic LMM evaluation benchmark (see Tab.~\ref{tab:ours_vs_others})  to rigorously evaluate and study LMMs for Arabic. 

To address the aforementioned issue, we introduce the first comprehensive Arabic LMM evaluation benchmark, named CAMEL-Bench. 
CAMEL-Bench is designed to encompass a wide range of tasks and focus on the Arabic-speaking population. It spans eight diverse domains and 38 sub-domains (see Fig. \ref{fig1_hierarchy}). 
The eight domains are: \textit{Multimodal understanding and reasoning}, \textit{OCR and document understanding}, \textit{chart and diagram understanding}, \textit{video understanding}, \textit{cultural-specific understanding}, \textit{medical image understanding}, \textit{Agricultural image understanding}, and \textit{remote sensing understanding}. 
Further, the 38 sub-domains (see Fig. \ref{fig1_hierarchy}) covered by our CAMEL-Bench are: visual understanding and reasoning, object hallucination evaluation, math and logic reasoning, scientific reasoning, VQA, infographics VQA, complex visual perception, real-world spatial understanding, multi-image understanding, object-level perception, newsletter, powerpoint slides, scene text, handwriting, lines, books, documents, charts, diagrams, tables, general video scenes, cultural-specific occasions, countries and landmarks in videos, countries and landmarks in images, food, celebrities, cultural VQA, basic medical science, clinical medicine, public health, pharmacy, diagnosis, medical understanding, plant types, fruit and veggies identification, plant illness, and geospatial imagery subdomains (land, transportation and construction).

Our CAMEL-Bench comprises 29,036 questions (see Fig. \ref{fig:qual_samples}) and follows an extensive manual verification process by native-speakers to ensure the resulting benchmark is of high-quality. We conduct extensive experiments using open and closed-source LMMs. Our results reveal the need for substantial improvement in handling of Arabic multimodal data, shedding light on the areas requiring further Arabic LMM improvements.

\section{CAMEL-Bench}


\definecolor{verylightgray}{rgb}{0.97, 0.97, 0.97}

\begin{table}[t]
    \centering
    \setlength{\tabcolsep}{3pt}
    \resizebox{0.47\textwidth}{!}{%
    \begin{tabular}{lcccccc}
    \hline
    \textbf{Domain/Characteristics} & \textbf{Exams-V*} & \textbf{CVQA*} & \textbf{Henna} & \textbf{KHATT} & \textbf{CAMEL-Bench}\\     &  & &  &  &  \small\emph{(ours)} \\
    \hline
    Multimodal Und. \& Reasoning & \cmark & \xmark & \cmark & \xmark & \cmark \\
    OCR \& Docs Und. & \xmark & \xmark & \xmark & \cmark & \cmark \\
    Charts \& Diagrams Und. & \cmark & \xmark & \xmark & \xmark & \cmark \\
    Video Und. & \xmark & \xmark & \xmark & \xmark & \cmark \\
     Medical Image Und. & \xmark & \xmark & \xmark & \xmark & \cmark \\
    Agricultural Image Und. & \xmark & \xmark & \xmark & \xmark & \cmark \\
    Remote-Sensing Und. & \xmark & \xmark & \xmark & \xmark & \cmark \\
    Cultural-Specific Und. & \xmark & \cmark & \cmark & \xmark & \cmark \\
    Open Source & \cmark & \cmark & \xmark & \cmark & \cmark \\
    \hline
     \rowcolor{verylightgray}
    Question Numbers & 823 & 200 & 1.1K & 5K & 29K \\
    \bottomrule
    \end{tabular}}

    \caption{Comparison of our CAMEL-Bench with existing Arabic LMM benchmarks: Exams-V \cite{das2024exams}, CVQA \cite{romero2024cvqa}, Henna\cite{alwajih2024peacock}, and KHATT \cite{mahmoud2014khatt}. Here * denotes that only Arabic part of benchmark is counted.}
    \label{tab:ours_vs_others}
    \vspace{-1em}
\end{table}

\begin{table*}[t!]
    \centering
    \setlength{\tabcolsep}{5pt}
    \resizebox{\textwidth}{!}{%
    \begin{tabular}{lllc}
    \hline
     \rowcolor{verylightgray}
    \textbf{Domains} & \textbf{Sub-Domains} & \textbf{Source} & \textbf{Number of Questions} \\
    \hline
    \multirow{10}{*}{Multimodal Understanding and Reasoning} 
    & Visual Understanding/ Reasoning & MME, MMBench, MMT-Bench-MI, SEED, MMMU & 3,971 \\
    & Object Hallucination Evaluation & CountBench, MMT-Bench-MI, POPE & 997 \\
    & Math and Logic Reasoning & MathVista & 531 \\
    & Scientific Reasoning & ScienceQA-IMG, Exams-V & 1,624 \\
    & Visual Question Answering & GQA, VizWiz, VQAv2 & 3,840 \\
    & InforGrahpics VQA & AI-Generated (GPT-4o), Pinterest & 120 \\
    & Complex Visual Perception & BLINK & 1,422 \\
    & Real-world Spatial Understanding & RealWorldQA & 624 \\
    & Multi-image Understanding & MMT-Bench-MI, MuirBench & 1,062 \\
    & Object-level Perception & COCO, ImageNet, Mocheg, Snli-Ve & 60 \\
    \hline
    \multirow{9}{*}{OCR and Document Understanding} 
    & Scanned Documents (OCR) & ArabicDatasetOCR & 480 \\
    & Scanned Documents (VQA) & MTVQA & 703 \\
    & Scene Text (OCR) & EvArEST & 1,217 \\
    & Books (OCR)& Historical Arabic Handwritten Text Recognition Dataset & 40 \\
    & PowerPoint Slides (OCR) & ISI-PPT-Dataset & 2,354 \\
    & PowerPoint Slides (VQA) & ISI-PPT-Dataset & 711 \\
    & Handwriting (OCR) & KHATT Line & 1,400 \\
    & Newsletters (OCR)& PATD & 506 \\
    & Lines (OCR) & PATS-01 & 520 \\
    \hline
    \multirow{3}{*}{Chart and Diagram Understanding} 
    & Charts & ChartQA & 745 \\
    & Diagrams Understanding & MMMU (diagrams), ICON-QA, AI-Generated, Pinterest, BCE-Arabic & 1,994 \\
    & Tables & BCE-Arabic, Excel & 81 \\
    \hline
    \multirow{3}{*}{Video Understanding} 
    & Countries/ Landmarks & Pexel & 87 \\
    & Cultural-Specific Occasions & Pexel & 24 \\
    & General Video Scenes& Video-MME & 654 \\
    \hline
    \multirow{3}{*}{Cultural Specific Understanding} 
    & Celebrities & arab-celeb-dataset & 444 \\
    & Food & arabic-food-101, Pexel & 347 \\
    & Countries/ Landmarks & Pexel & 494 \\
    \hline
    \multirow{8}{*}{Medical Imaging Understanding} 
    & Basic Medical Science & MMMU, MMMU Pro & 89 \\
    & Clinical Medicine &MMMU, MMMU Pro & 83 \\
    & Public Health & MMMU, MMMU Pro & 87 \\
    & Pharmacy & MMMU, MMMU Pro & 82 \\
    & Diagnosis & MMMU, MMMU Pro& 87 \\
    & Medical Understanding & MMT-MI-Bench & 78 \\
    \hline
    \multirow{1}{*}{Agricultural Image Understanding} 
    & Agriculture Image Understanding & AgroGPT & 769 \\
    \hline
    \multirow{1}{*}{Remote Sensing Understanding} 
    & Remote Sensing Understanding & GeoChat & 709 \\
    \hline
     \rowcolor{verylightgray}
    \textbf{Total} & & & \textbf{29,036} \\
    \hline
    \end{tabular}}
    \caption{Different data sources used for 38 sub-domains corresponding to eight domains, with around 29k questions in total. The different data sources include: MME \cite{fu2023mme}, MMBench \cite{MMBench}, MMT-Bench-MI \cite{ying2024mmtbench}, SEED \cite{li2024seed}, MMMU \cite{yue2023mmmu}, MMMU-Pro \cite{yue2024mmmu}, CountBench \cite{paiss2023teaching},  POPE \cite{Li-hallucination-2023}, MathVista \cite{lu2023mathvista}, Exams-V (Arabic portion) \cite{das2024exams}, ScienceQA-IMG \cite{lu2022learn}, GQA \cite{hudson2019gqa}, VizWiz \cite{bigham2010vizwiz}, VQAv2 \cite{goyal2017making}, BLINK \cite{fu2024blink}, MuirBench \cite{wang2024muirbench}, COCO \cite{lin2014microsoft}, Imagenet \cite{deng2009imagenet}, Mocheg \cite{yao2023end}, Snli-Ve \cite{xie2019visual},  Pinterest \cite{Pinterest2024}, RealWorldQA \cite{xai2024grok},  PATS-01 \cite{al2010arabic}, KHATT \cite{mahmoud2014khatt}, PATD \cite{PATD}, Historical Arabic Handwritten Text Recognition Dataset \cite{najam2024historical}, ISI-PPT-Dataset \cite{wu2017iccv}, EvArEST \cite{hassan2021arabic}, MTVQA \cite{tang2024mtvqa}, ChartQA \cite{masry-etal-2022-chartqa}, IconQA \cite{lu2021iconqa}, BEC-Arabic \cite{saad2016bce}, Claude-3.5 \cite{claude}, arab-celeb-dataset \cite{mohammadalfaifi_arabceleb}, arabic-food-101 \cite{arabicfood101}, Countries and landmarks \cite{Wikipedia2024, Pexel2024, youtube}, Pexel \cite{Pexel2024}, AgroGPT \cite{awais2024agroGPT}, GeoChat \cite{kuckreja2023geochat}. These data sources are carefully translated and verified to ensure quality and relevance. }
    \label{tab:overview}
\end{table*}

\subsection{Data Collection}
Our dataset encompasses eight diverse domains to ensure a versatile multi-task Arabic LMM benchmark for different real-world scenarios. Each domain is further sub-divided into different sub-domains, each focusing on a distinct aspect. During the data collection process, we either utilize available Arabic multimodal data samples or employ samples from existing English-centric LMM benchmarks. These English samples are then translated to Arabic via GPT-4o and verified. Alternatively, we manually collect and generate the Arabic samples for remaining sub-domains from internet. Tab.~\ref{tab:overview} presents the details of different data sources used for data collection for the 38 sub-domains corresponding to eight domains, with around 29k questions in total.

 \begin{figure*}[!ht]
    \centering
      \includegraphics[width=\textwidth]{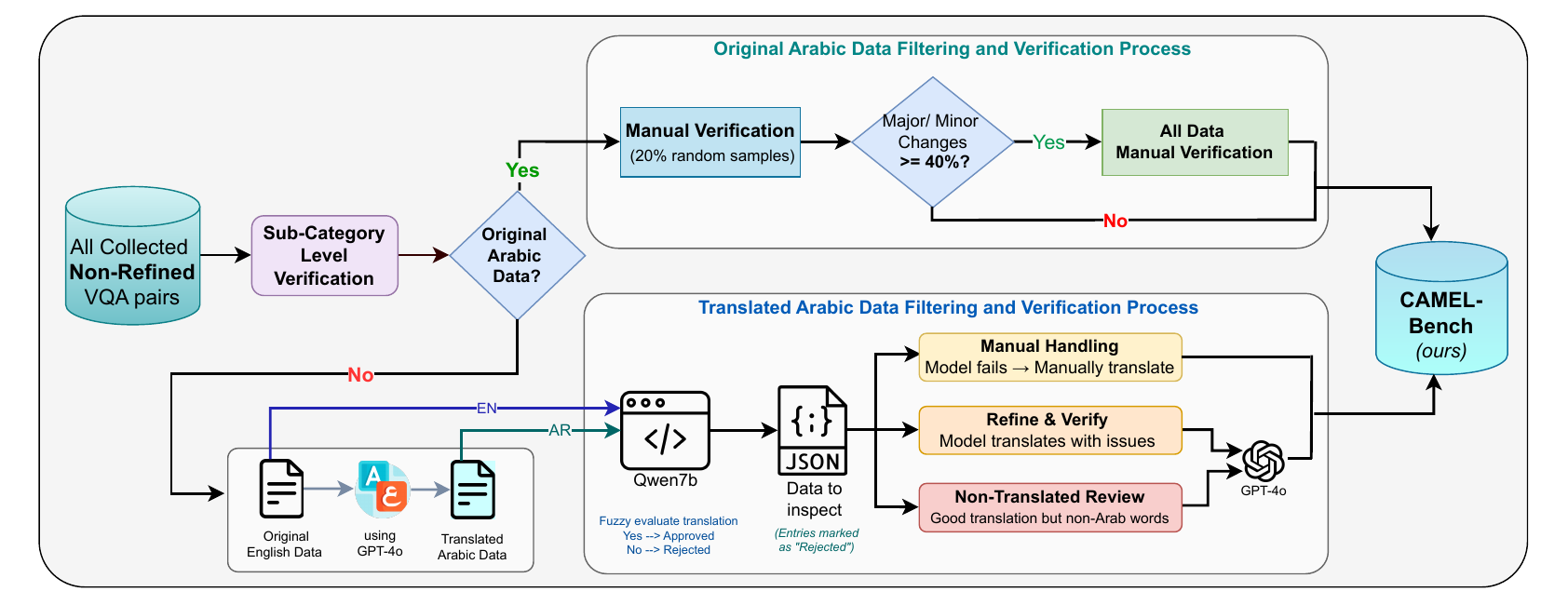}
        \caption{The CAMEL-Bench Filtering and Verification Pipeline consists of two paths: Original Arabic and translated Arabic. For original Arabic (top row), a 20\% random sample undergoes manual verification; if errors are below 40\%, the data passes; otherwise, the entire sub-category is reviewed. For Translated Arabic (bottom row), We employ Qwen7B model \cite{qwen} to assess semantic similarity between the original and translated question-answer pairs on fuzzy-basis evaluation. Pairs passing the evaluation proceed, while those that fail undergo manual review. Based on this, data may require \emph{Manual Handling} for manual re-translation, \emph{Refine \& Verify} for refinement through the model, or \emph{Non-Translated Review} where the data is re-sent for translation due to the absence of an Arabic version.}
        \label{fig2_pipeline}
        \vspace{-1em}
\end{figure*}

\subsection{Question-Answers Pairs Generation}
We note that a major part of our original Arabic data is not derived from ready-made VQA datasets. Some sub-domains, such as celebrities and food, consist of image-only data, while others, like Pexel’s countries and landmarks, contain image-caption pairs. To create a rich and diverse VQA corpus, we first ensure that each image is accompanied by detailed contextual information. This context is sourced from a combination of Wikipedia (e.g., for food-related data), manual curation (e.g., for countries and landmarks in videos), and AI-generated content based on a manually provided context (e.g., for diagrams and infographics).
Next, we generate multiple-choice questions (MCQs) for each sample using the GPT-4o model. The prompt is meticulously crafted to adhere to key criteria: each sample generates three multiple-choice questions (MCQs), with four distinct, non-synonymous options per question, only one of which is correct. The questions contain no embedded hints, ensuring that answers are derived exclusively from the image, without requiring prior knowledge. Additionally, the image must provide enough information to fully support the correct answer, eliminating the need for guesswork.
In total, this process produces a corpus of 4.4K generated questions with 17.7K answers, enabling a comprehensive set of questions for evaluation.

\subsection{Data Filtering and Verification}
The data collection and question-answer pair generation process lead to over 41k questions in total which then undergoes to filtering and verification process. The CAMEL-Bench filtering and verification process (see Fig.~\ref{fig2_pipeline}) is carefully conducted based on whether the QA text is originally Arabic or translated into Arabic from English language. For all sub-domains derived from original Arabic context, we take a 20\% randomly sampled subset for manual verification. In case if the error remains less below a 40\% threshold, the sub-category is accepted into CAMEL-Bench. Alternatively, the entire sub-category undergoes manual review.

In case of the translated Arabic data from English, the original English context is also incorporated into the filtering and verification process. Here, Qwen7B \cite{qwen} is used to compare the semantic similarity between the English and the English-translated data at the QA-pair level using fuzzy evaluation. To ensure the model understands semantic similarity in Arabic, we provided 5 few-shots prompting. Subsequently, QA-pairs rejected by Qwen7B \cite{qwen} are manually reviewed, resulting in one of three outcomes. \emph{Manual Handling} implying that data requires full re-translation. 
 \emph{Refine and Verify} referring that the translation can be refined using the model. \emph{Non-Translated Review} implying that the non-translated data is re-sent to the model for translation. Consequently, we obtain 29,036 high-quality questions.

\begin{table*}[t!]
    \centering
    \setlength{\tabcolsep}{5pt}
    \resizebox{\textwidth}{!}{%
    \begin{tabular}{lcccccccc}
    \toprule
    \multirow{2}{*}{ Method } & MM Understanding & OCR \& Document & Charts \& Diagram & Video & Cultural Specific & Medical & Agro & Remote Sensing \\
    & \& Reasoning & Understanding & Understanding & Understanding & Understanding & Imaging & Specific & Understanding \\
    \midrule
    GPT-4o & 57.90 & 59.11 & 73.57 & 74.27 & 80.86 & 49.90 & 80.75 & 22.85 \\
    GPT-4o-mini & 48.82 & 42.89 & 64.98 & 68.11 & 65.92 & 47.37 & 79.58 & 16.93 \\
    Gemini-1.5-Pro & 46.67 & 36.59 & 47.06 & 42.94 & 56.24 & 33.77 & 72.12 & 17.07 \\
    Gemini-1.5-Flash & 45.58 & 33.59 & 48.25 & 53.31 & 46.54 & 42.86 & 76.06 & 14.95 \\
    Pangea-7B & 40.09 & 26.47 & 38.87 & 49.01 & 20.34 & 31.99 & 74.51 & 6.67 \\
    Qwen2-VL-2B & 40.59 & 25.68 & 27.83 & 38.90 & 34.27 & 29.12 & 52.02 & 12.56 \\
    InternVL2-8B & 30.41 & 15.91 & 30.27 & 51.42 & 20.88 & 29.48 & 44.47 & 5.36 \\
    LLaVa-NeXt-7B & 26.33 & 19.12 & 27.56 & 44.90 & 28.30 & 22.54 & 42.00 & 8.33 \\
    \bottomrule
    \end{tabular}}
    \caption{\textbf{Performance comparison of different closed-and open-source LMMs on CAMEL-Bench.} We present per-domain results of seven LMMs: GPT-4o~\cite{gpt4o}, GPT-4o-mini~\cite{gpt4o}, Gemini-1.5-Pro~\cite{gemini_announcement}, Gemini-1.5-Flash~\cite{gemini_announcement}, Pangea-7B~\cite{pangea}, Qwen2-VL~\cite{bai2023qwen}, InternVL2-8B~\cite{chen2023internvl}, and LLaVaNeXt-7B~\cite{liu2024llavanext}. 
    GPT-4o excels in most domains,  while GPT-4o-mini offers an impressive balance of performance and model size.  All models struggle with remote sensing, medical imaging, OCR \& document understanding, and general multimodal understanding and reasoning domains. Open-source models like InternVL2-8B and LLaVaNeXt-7B show a decline in performance across domains, with their best results in video understanding.
    }
    \label{tab:methods_comparison}
    
\end{table*}

\begin{figure*}[!t]
    \centering
      \includegraphics[width=1\textwidth]{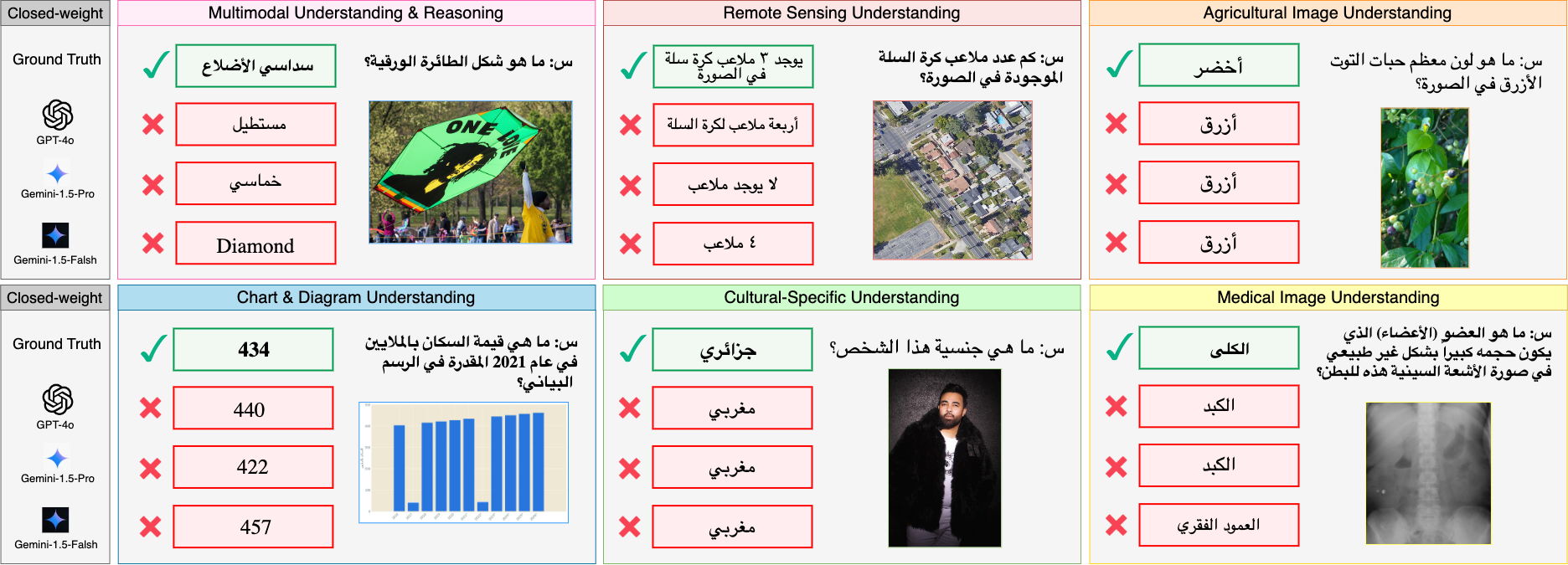}
        \caption{Qualitative example highlighting different scenarios where different closed-weight models struggle on CAMEL-Bench. The correct response is shown in green, and the incorrect one in the red box.
  }
        \label{fig:eval_samples}
        \vspace{-1em}
\end{figure*}

\section{CAMEL-Bench Benchmark Evaluation}

\noindent \textbf{Evaluation Metrics:} Our evaluation framework is designed with three specialized metrics, each carefully aligned to different types of datasets and tasks. For MCQ datasets like MMT~\cite{ying2024mmtbench} and MMMU~\cite{yue2023mmmu}, we utilize exact match accuracy to ensure precise evaluation. For optical character recognition (OCR) datasets, such as PATS~\cite{al2010arabic} and Evarest~\cite{hassan2021arabic}, where accurate text extraction is critical, we adopt edit distance~\cite{ristad1998learning} as the key metric. For more flexible datasets like VQAv2~\cite{goyal2017making}, MathVista~\cite{lu2023mathvista}, and GeoChat~\cite{kuckreja2023geochat}, where multiple synonymous answers can be considered correct. we implement a fuzzy evaluation method for all such datasets. This approach uses GPT-4o to compare the predicted answer with the ground truth, while accounting for the context of the question. By incorporating these diverse metrics, our evaluation provides a robust and comprehensive assessment that adapts to the unique demands and response formats of each dataset.

\begin{figure*}[!t]
    \centering
      \includegraphics[width=1\textwidth]{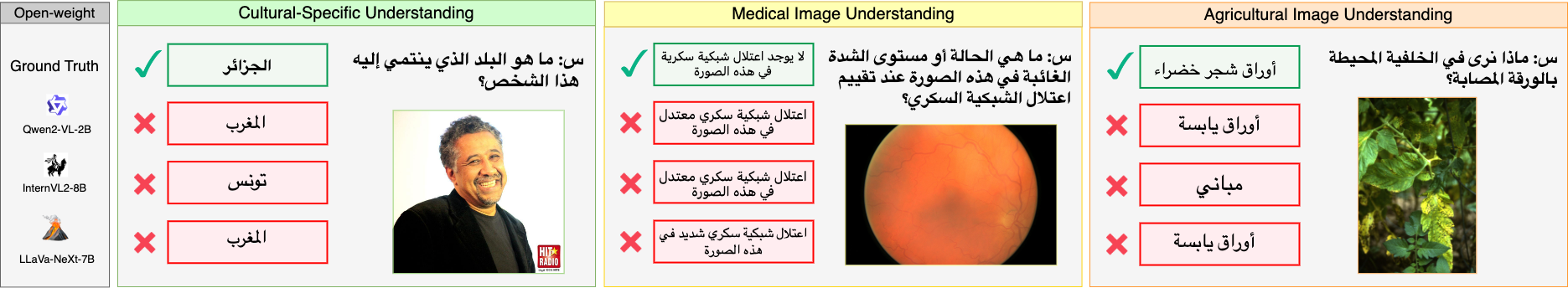}
        \caption{Qualitative example highlighting different scenarios where different open-weight models struggle on CAMEL-Bench. The correct response is shown in green, and the incorrect one in the red box.}
        \label{fig:open_samples}
        \vspace{-1em}
\end{figure*}

Tab.~\ref{tab:methods_comparison} presents a comparative evaluation of five different models on a range of multimodal (MM) understanding tasks, each assessing the capabilities of the models in distinct domains. The models include GPT-4o, GPT-4o-mini, Gemini-1.5-Pro, Gemini-1.5-Flash, and Qwen2-VL-2B, evaluated on key tasks such as multimodal reasoning, OCR \& document understanding, chart \& diagram interpretation, video analysis, and several domain-specific tasks like cultural understanding, medical imaging, agricultural (agro) understanding, and remote sensing. GPT-4o excels across tasks, leading in MM reasoning (57.90), chart/diagram understanding (73.57), video analysis (74.27), cultural (80.86) and agro-specific understanding (80.75).
Models perform well on MCQs and binary-option tasks due to guessing probability and context. Infographics, designed for easy interpretation, also see high accuracy across all models.
In contrast, Arabic OCR tasks, particularly in datasets like Khatt, historical documents prove exceptionally challenging. This difficulty stems from the complex nature of Arabic script, which uses ligatures and diacritics (small markings that alter pronunciation and meaning). Remote sensing understanding also remains difficult, with scores like 22.85 (GPT-4o) and 16.93 (Qwen2-VL-2B), highlighting the complexities of interpreting satellite imagery. 

Among the open-source models evaluated on our Arabic multimodal benchmark, Pangea-7B stands out by outperforming InternVL2-8B and LLaVaNeXt-7B in key areas. It achieves higher scores in multimodal understanding and reasoning (40.09), OCR and document understanding (26.47), and charts and diagram understanding (38.87). This suggests that Pangea-7B's multilingual and culturally diverse training data enhance its ability to handle complex tasks across different languages and cultures. However, similar to other open-source models, Pangea-7B struggles in remote sensing understanding, scoring 6.67, highlighting challenges with specialized tasks. Overall, Pangea-7B's performance underscores the benefits of incorporating diverse linguistic and cultural data in training multilingual multimodal LLMs while indicating areas for improvement.

The Fig.~\ref{fig:eval_samples} and Fig.~\ref{fig:open_samples} highlight a critical challenge in Arabic multimodal understanding, where all models fail to accurately interpret the linguistic context in the provided CAMEL-Bench samples. 
This underscores the complexity of Arabic linguistics, especially in multimodal tasks, and the need for more robust language models that can effectively integrate both visual and textual information in Arabic contexts.

\section{Conclusion, Limitations and Societal Impact}
We present a comprehensive and diverse benchmark, named CAMEL-Bench, for Arabic LMM evaluation. To the best of our knowledge, CAMEL-Bench is the first comprehensive Arabic LMM benchmark comprising eight diverse domains and 38 sub-domains with around 29k questions that are filtered from a larger pool of 41k samples with the quality verified by native speakers. We conduct extensive evaluations of open- and closed-source LMMs, highlighting the need for substantial improvements in different areas for future Arabic LMM development.
Although our CAMEL-Bench strives to significantly contribute towards developing sophisticated Arabic LMMs, we note that it mainly covers modern standard Arabic and does not fully explore other Arabic dialects. As the data samples are either based on existing datasets or new data that is crawled from the internet, it is possible that CAMEL-Bench exhibits biases already existing in the benchmarks. Nevertheless, we believe CAMEL-Bench is a step towards the inclusion of Arabic language and Arabic-speaking populations in accessing the benefits of LMMs.

\section*{Acknowledgements}
We sincerely thank Farah Husain Salem Abdullah Alharth for her valuable contributions to the manual data verification process.
{
\small
\bibliographystyle{ieeenat_fullname}
\bibliography{custom}
}

\clearpage 
\appendix
\section{Appendix}
\label{sec:appendix}

\section{More on Dataset Curation}
The dataset utilized in this work was carefully curated with a rigorous focus on data quality, relevance, and diversity. Our curation process involved selecting multimodal data from various domains, including images, text, videos, and specialized fields such as medical imaging, agriculture, and remote sensing. To ensure the integrity and accuracy of the dataset, we employed multiple stages of data verification. This process involved cross-validation, thorough verification procedures for Arabic content, and the integration of standardized data sources where applicable.

\section{Dataset Overview and Task Splits}
This section provides a comprehensive breakdown of the datasets used across eight distinct categories, illustrating the diversity and depth of our evaluation framework. Each category is further divided into sub-domains, ensuring that the multimodal models are rigorously tested on a wide range of tasks and datasets. 
This structure guarantees comprehensive coverage and introduces varied challenges to thoroughly assess model performance.
Refer to Tab.~\ref{tab:overview} for a detailed breakdown of the data categories with their statistics.

\subsection{Multimodal Understanding and Reasoning}
This category encompasses various sub-domains such as visual understanding, object hallucination evaluation, and complex visual perception. Key datasets include MME, MMBench, ScienceQA-IMG, and VQA2. These datasets test the model's ability to handle intricate reasoning tasks across both visual and textual inputs, with a total of 3,971 questions under the visual understanding sub-domain, and significant representation from other tasks like scientific reasoning (1,624 questions) and object-level perception (60 questions).

\subsection{OCR and Document Understanding}
Document understanding covers scanned documents, scene understanding, text extraction, and more. This category emphasizes precise OCR and textual recognition from images and scanned materials. Datasets like ArabicDatasetOCR and ISI-PPT-Dataset challenge the model to process a diverse range of document types. A substantial number of questions come from Handwritten Text datasets (1,400 questions) and PPT OCR (2,354 questions), ensuring the model is evaluated across both structured and unstructured document types.

\subsection{Chart and Diagram Understanding}
In chart and diagram interpretation, models are tested on understanding visual representations of data, such as charts, diagrams, and tables. This includes datasets like ChartQA, MMMU, and BCE-Arabic. The evaluation focuses on tasks such as understanding diagrammatic reasoning and tabular data with 1,994 questions from diagram datasets and 745 questions involving charts, providing a robust examination of the model's ability to interpret visual data efficiently.

\subsection{Video Understanding}
This category assesses the model's ability to process and comprehend video data, focusing on tasks like recognizing countries, landmarks, and occasions. Video-MME is a prominent dataset, contributing 654 questions to the evaluation. The inclusion of diverse sub-domains, such as recognizing cultural aspects through video, highlights the importance of temporal and visual information synthesis in multimodal reasoning.

\subsection{Cultural Specific Understanding}
The cultural understanding domain tests the model's capacity to handle tasks specific to certain cultures, including food, landmarks, and celebrities. Datasets like arabic-food-101 and Pexel challenge the model to recognize culturally significant items, with 444 questions focused on celebrities and 494 on countries/landmarks. These tasks highlight the model’s ability to adapt and generalize across different cultural contexts.

\subsection{Medical Imaging}
Covering a range of sub-domains in the medical field, this category includes tasks related to basic medical science, clinical medicine, and public health, using datasets like MMMU and MMT-MI-Bench. These datasets assess the model's potential in specialized medical contexts, with over 1,200 questions spanning across diagnosis, medical understanding, and pharmacy, ensuring a rigorous evaluation of the model’s performance in handling critical medical information.

\subsection{Agricultural Image Understanding}
The agricultural domain is represented through datasets like AgroGPT, with 769 questions focused on agricultural understanding tasks. These tasks test the model’s capacity to process and interpret images related to agricultural settings, reinforcing the model’s ability to work with real-world scenarios in agriculture and environment-based challenges.

\subsection{Remote Sensing Understanding}
This category evaluates the model's ability to handle remote sensing data, specifically focusing on geographical data interpretation through datasets like GeoData VQA and GeoChat. With 709 questions in this domain, the model is tested on its spatial reasoning and understanding of complex remote-sensing imagery, crucial for applications in fields like environmental monitoring and geography.

In total, the dataset includes 29,036 questions across all categories, providing a comprehensive and diverse benchmark for evaluating the multimodal model's performance across a wide spectrum of tasks. This balanced distribution ensures that the model is tested thoroughly, with each domain offering unique challenges and insights into the model’s strengths and areas for improvement.

\section{CAMEL-Bench Data Samples}
Fig.~\ref{fig:qual_samples} showcases CAMEL-Bench's versatility across eight distinct domains, covering tasks like Multimodal Reasoning, OCR \& Document Understanding, Chart \& Diagram Interpretation, Video Scene Analysis, and more specialized areas like Remote Sensing, Agricultural Image Analysis, Medical Image Interpretation, and Cultural-Specific Knowledge. Each domain presents unique challenges, from logical reasoning and handwritten text recognition to medical diagnostics and cultural symbol identification. This variety emphasizes CAMEL-Bench's strength in supporting the development of AI systems capable of addressing real-world applications in healthcare, agriculture, geospatial analysis, and cross-cultural contexts.

\end{document}